\documentclass[10pt,letterpaper]{article}

\usepackage{cogsci}
\usepackage{pslatex}
\usepackage{apacite}

\usepackage{natbib}
\usepackage[dvipsnames]{xcolor}
\usepackage{graphicx}
\graphicspath{{./figs/}}
\usepackage{wrapfig}
\usepackage{amsmath}
\usepackage{float}
\usepackage{flushend}

\title{Learning a face space for experiments on human identity}
 
\author{
    {\bf Jordan W. Suchow* (suchow@berkeley.edu)} \\
    {\bf Joshua C. Peterson* (jpeterson@berkeley.edu)} \\
    {\bf Thomas L. Griffiths (tom\_griffiths@berkeley.edu)} \\
  Department of Psychology\\ University of California, Berkeley}

\begin{document}

\maketitle

\begin{abstract}

Generative models of human identity and appearance have broad applicability to behavioral science and technology, but the exquisite sensitivity of human face perception means that their utility hinges on the alignment of the model's representation to human psychological representations and the photorealism of the generated images. Meeting these requirements is an exacting task, and existing models of human identity and appearance are often unworkably abstract, artificial, uncanny, or biased. Here, we use a variational autoencoder with an autoregressive decoder to learn a face space from a uniquely diverse dataset of portraits that control much of the variation irrelevant to human identity and appearance. Our method generates photorealistic portraits of fictive identities with a smooth, navigable latent space. We validate our model's alignment with human sensitivities by introducing a psychophysical Turing test for images, which humans mostly fail. Lastly, we demonstrate an initial application of our model to the problem of fast search in mental space to obtain detailed ``police sketches'' in a small number of trials.

\end{abstract}

\let\thefootnote\relax\footnotetext{\noindent *JWS and JCP contributed equally to this submission.}

\section{Introduction}

Generative models of human identity and appearance have broad applicability to behavioral science and technology. For example, psychologists and neuroscientists can use them to create experimental stimuli, varying the degree of similarity between a target face and a lineup of other faces to test an observer's face recognition abilities. Computer vision researchers can use them to create face-detection and face-recognition technologies for applications in security and social media. And artists can use them to create engaging animated characters in video games.

Any generative model of human identity and appearance can be cleaved into two parts: (1) a so-called latent ``face space'', the underlying  multidimensional psychological space of perceived features and properties that specifies which faces are similar to which other faces, and (2) a renderer that converts a point in the face space to an image. The success of a generative model of human identity and appearance, then, depends on the quality of both the face space and the renderer. 

One way to create a high-quality generative model of human identity and appearance is to learn the model directly from a corpus of portraits, using machine learning methods such as deep neural networks to construct a latent face space and renderer that reflect the training data. With an adequately large and diverse data set that contains variation relevant to human identity and appearance, such an approach holds considerable promise in achieving the goal of creating generative models with smoothly navigable and interpretable face spaces that render into photorealistic portraits representative of human diversity.

\begin{figure}
    \centering
    \includegraphics[width=0.9\linewidth]{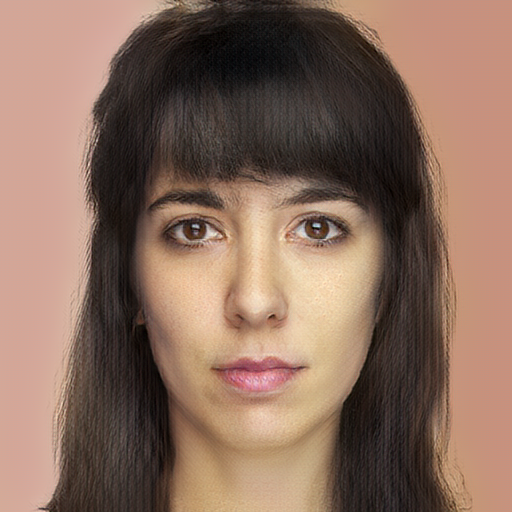}
    \vspace{-2mm}
    \caption{\label{242_ne8x}A fictive identity from the model  at $512\times512$ px.}
    \vspace{-5mm}
\end{figure}

The plan of the paper is as follows. We begin by presenting a uniquely diverse dataset of portraits that control much of the variation irrelevant to human identity and appearance. Next we train a deep neural network (specifically, a variational autoencoder with an autoregressive decoder) on these images. We then perform a visual Turing test for assessing the quality of such models for use with humans, and show that our best model nearly passes the test. Next, we show that samples from the model can be enlarged and drastically improved while preserving identity. Lastly, we demonstrate an initial application of our model to quickly extracting information from human mental representations, yielding expertless ``police sketching'', reducing by tenfold the number of human judgments required in comparison to previous methods .

\section{Learning a latent face space \nopagebreak\\ for human experiments}

The face space learned by a neural network reflects in part the images used to train it. Many image sets of faces used for training networks are confounded by variation unrelated to identity, while lacking representative variation essential to it. Confounded properties include those of the photographer and equipment (photographer, camera model, lens model, focal length, focal point, aperture, exposure time, distance to subject, camera placement, light design), the subject's ephemeral state (facial expression, pose, arousal), the environment (background imagery, ambient lighting), and post-processing (resolution, digital format, color grading, white balance, gamma correction, compression quality, watermarking, and digital alteration of skin complexion, hair color, and face shape), among others. Though learning a representation of identity that is invariant with respect to these properties requires a training set that varies along them, holding these properties constant greatly simplifies the learning problem in contexts where such invariance is irrelevant \citep{zhang2009face}. And though there are public datasets of human faces that control much of this variation, they tend to have too few unique individuals, precluding the possibility of learning a universal face space.

Because our goal is to produce a latent space of human identity useful for empirical experiments with humans, it is just as important that this representation be decodable into images that are free of distortions and artifacts, because human judgments of such images may often mistake them for features, or otherwise remove the natural context of the visual stimuli that humans often reason within. This is a difficult problem in machine learning, and will likely require a dataset with well-controlled variation so it can be learned using current methods, yet it also requires just the right diversity, avoiding heavy bias towards attractive celebrities and certain ethnicities.

\subsection{The Human\ae \, Dataset}

To assist in the effective training of a representative latent face space for human identity, we use a novel dataset based on Human{\ae}, an artistic work by Ang\`elica Dass that explores human diversity by mapping the gamut of skin tone across over $3,300$ people, with minimal constraints on participant selection. Of particular note is that the images are so nearly unvaried in their production that variation across the images focuses solely on the diversity of human visual identity. The abundance of controlled variation relevant to identity makes the dataset useful despite its relatively small size when compared to other face datasets used to train deep neural networks.

\begin{figure*}[!ht]
  \begin{center}
    \includegraphics[trim = 0mm 294mm 0mm 0mm, clip, width=1.0\textwidth]{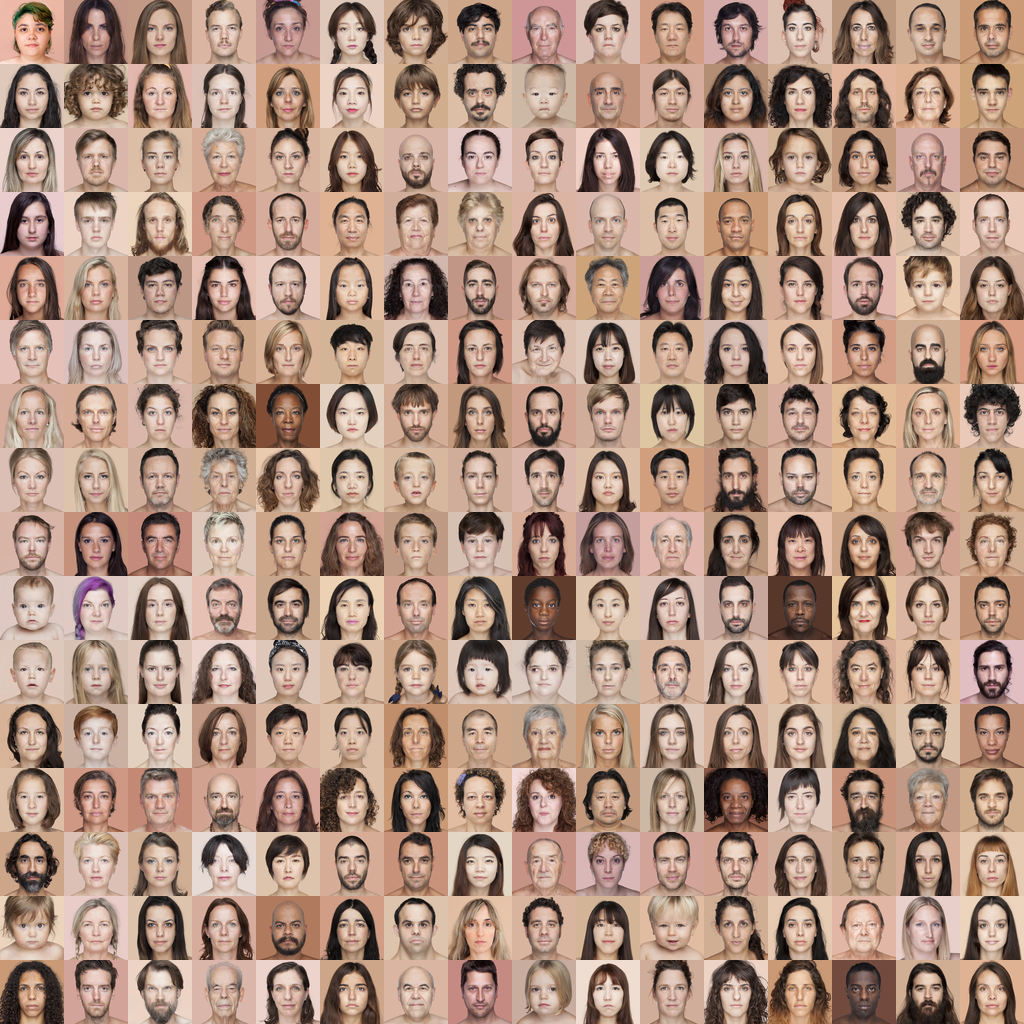}
  \vspace{-6mm}
  \caption{Sample images from the aligned/cropped Human{\ae} dataset.}
  \label{samples_humanae}
  \end{center}
\end{figure*}

The dataset consists of $3,353$ front-facing portraits scraped from the project's website (http://humanae.tumblr.com/) with permission. Portraits were first resized by Lanczos resampling to $1024\times1024$ pixels. Next, the portraits were aligned through Procrustes superimposition of facial landmarks, which rotates, scales, and translates each portrait to minimize the squared distances between the detected landmarks and those on a target, here a composite that places each landmark at its average position across all portraits in the set. Facial landmarks were detected using a pre-trained ensemble-of-regression-trees detector \citep{kazemi2014one}. After cropping a $640\times640$ square from the center of each portrait, they were then downsampled to the training size of $512\times512$ pixels by Lanczos resampling. 

We note that, although this dataset is small by the standards of modern machine learning applications, it is nevertheless effective given its well-controlled production by the artist. Forty-eight images from this dataset can be seen in Figure \ref{samples_humanae}. The background of each image was matched to the skin tone of each face by the artist, a property we left intact.

\subsection{Generative Models}

\begin{figure*}[!h]
  \begin{center}
    \includegraphics[trim = 0mm 294mm 0mm 0mm, clip, width=1.0\textwidth]{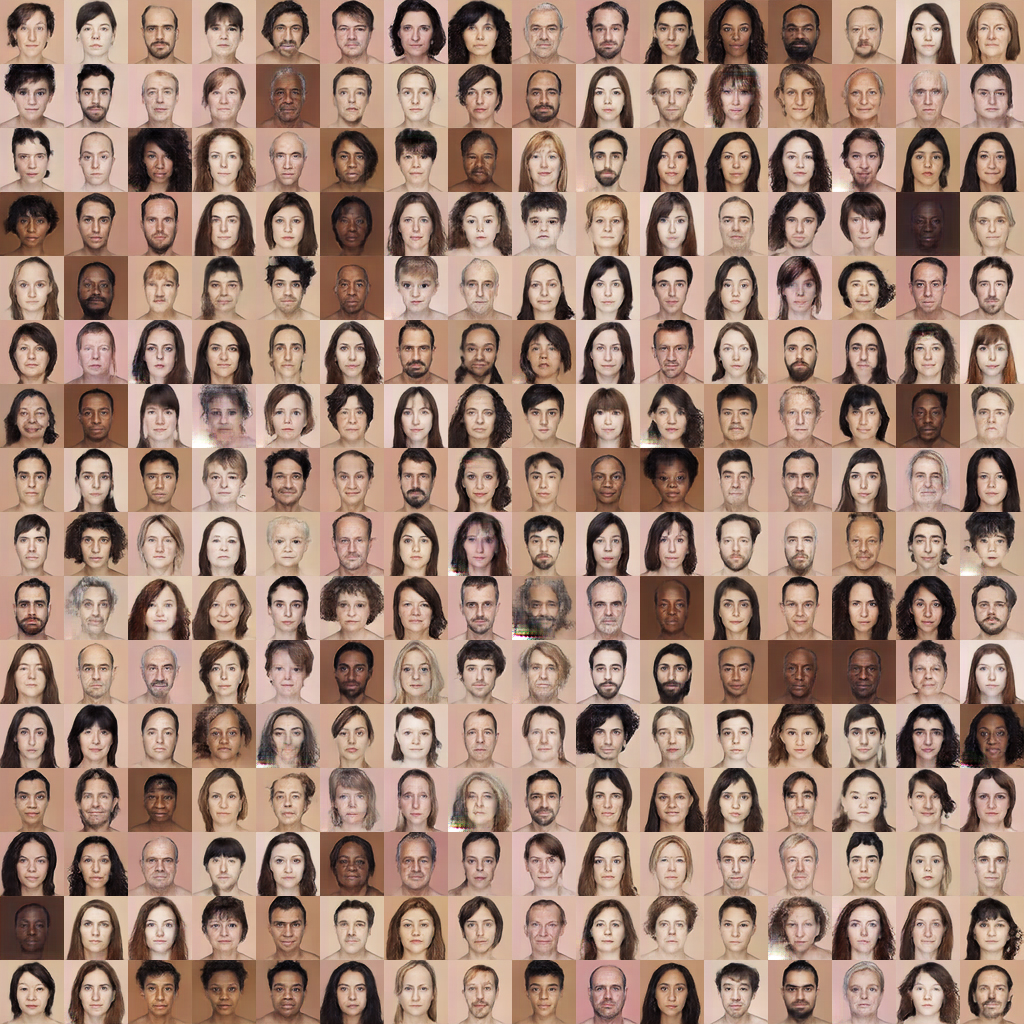}
  \vspace{-6mm}
  \caption{Samples from WGAN-GP trained on human\ae.}
  \label{samples_wgangp}
  \end{center}
\end{figure*}

To assess candidate face spaces, we employed two families of generative models and one hybrid. Generative Adversarial Networks (GANs; \citealp{goodfellow2014generative}) are a kind of deep neural network known to produce high-quality samples with reasonably good latent spaces. In particular, we use a type of GAN called WGAN-GP \citep{gulrajani2017improved} because it showed slight improvements over other GANs we tested. Samples from the best-performing model are shown in Figure \ref{samples_wgangp}. As expected, the quality is high, despite numerous artifacts. 

\begin{figure*}[!h]
  \begin{center}
    \includegraphics[trim = 0mm 294mm 0mm 0mm, clip, width=1.0\textwidth]{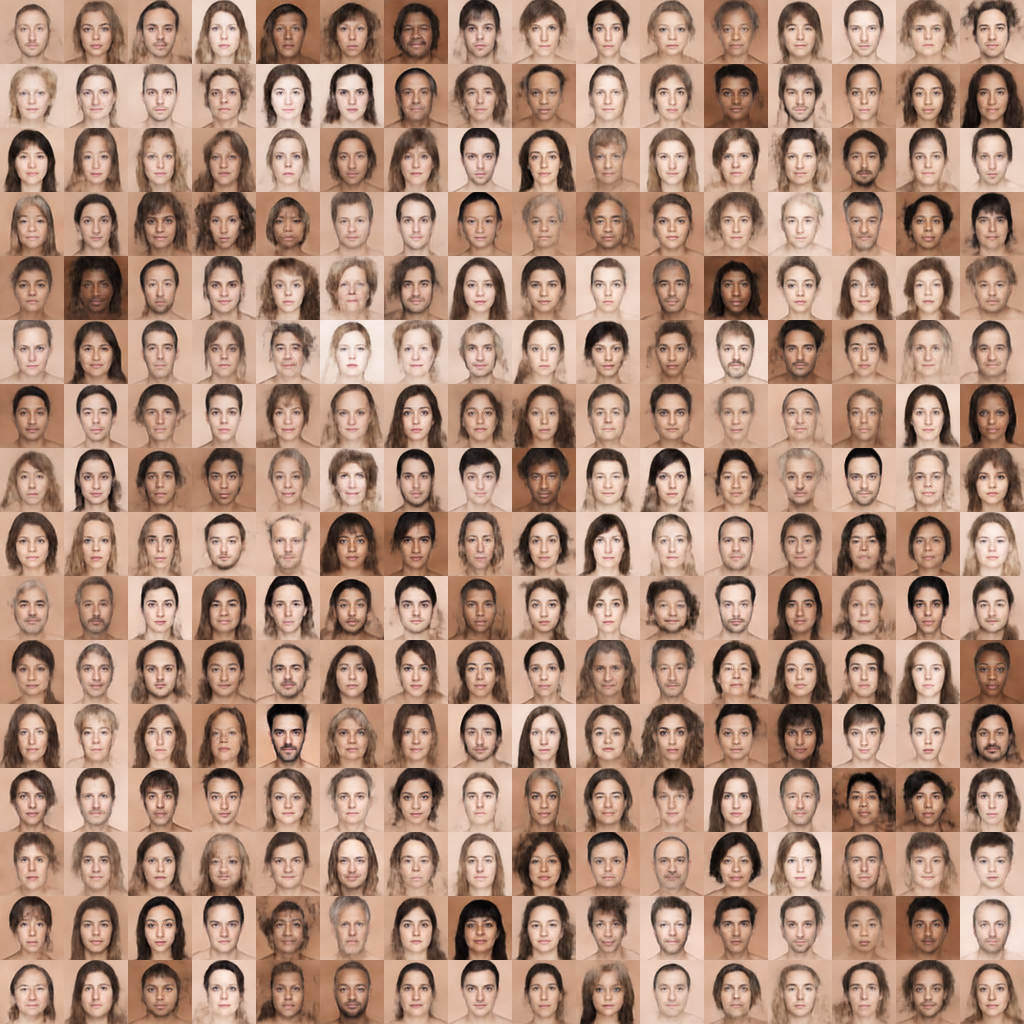}
  \vspace{-6mm}
  \caption{Samples from DFC-VAE trained on human\ae.}
  \label{samples_dfcvae}
  \end{center}
\end{figure*}

The second type of model that we explored are variational autoencoders (VAEs; \citealp{kingma2013auto}), another type of deep neural network. Most VAEs produce samples and reconstructions that are too smooth for our purposes, appearing slightly abstract and with a soft focus. To this end, we opted to train a deep ``feature-consistent'' variational autoencoder (DFC-VAE; \citealp{hou2017deep}), which preserves the spatial correlations between features, improving the appearance of the generated images. Surprisingly, samples from this model are indeed competitive with WGAN-GP (Figure \ref{samples_dfcvae}), although the model has difficulty in rendering hair.

\begin{figure*}[!h]
  \begin{center}
    \includegraphics[trim = 0mm 294mm 0mm 0mm, clip, width=1.0\textwidth]{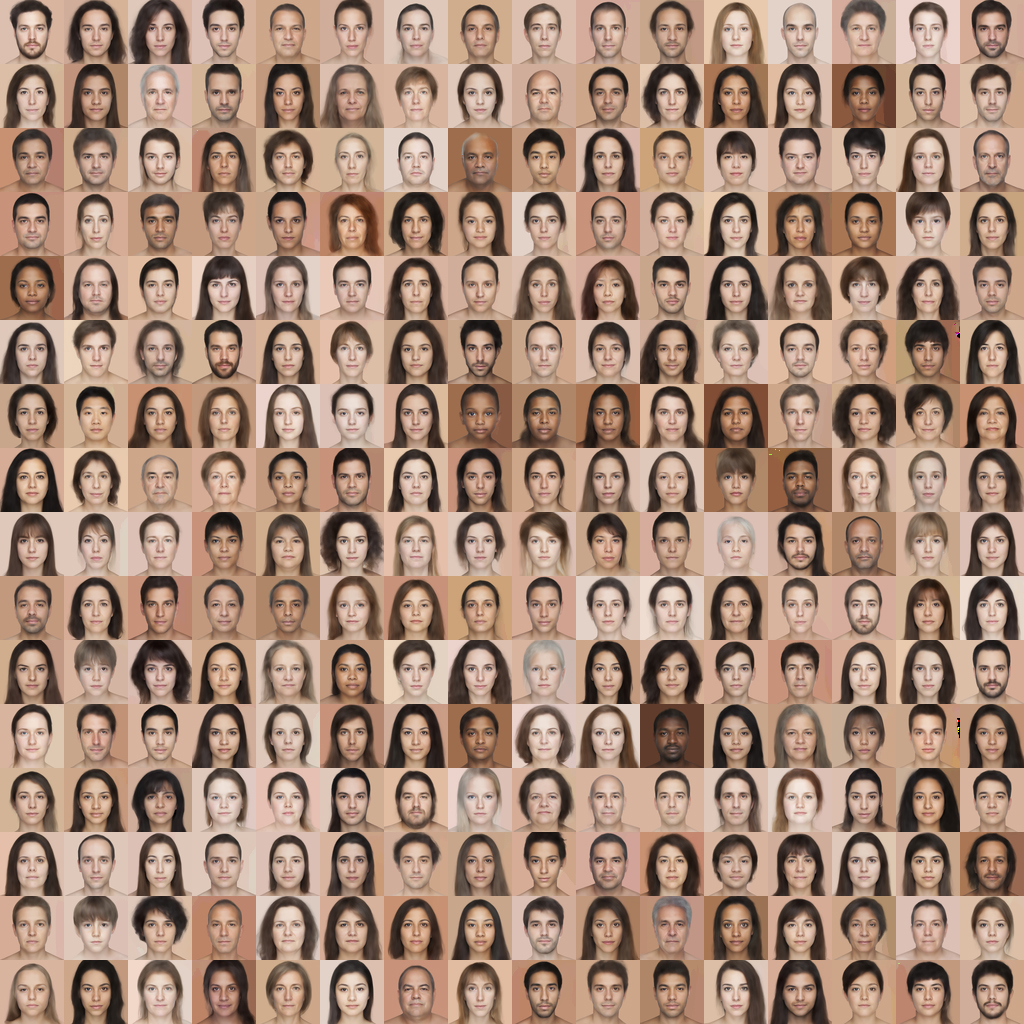}
  \vspace{-6mm}
  \caption{Samples from PixelVAE trained on human\ae.}
  \label{samples_pixelvae}
  \end{center}
  \vspace{-5mm}
\end{figure*}

Lastly, we opted not to train autoregressive models, which sample pixels sequentially, with each pixel depending on the value of all pixels previously sampled. Though these autoregressive models perform well, they do not provide a navigable latent space and thus are not suitable for our purposes. Instead, we trained a hybrid model, PixelVAE \citep{gulrajani2016pixelvae}, which provides a navigable face space that captures course variation, and uses a (partially) autoregressive decoder only to render finer visual detail. This is a particularly good match for capturing human identity from portraits because it allows for an accessible latent space, while encoding fine image detail elsewhere. Indeed, we find that this approach is effective for our data. The samples in Figure \ref{samples_pixelvae} appear perfect in many respects, including striking detail, variation, and little to no artifacts. Though it seems clear to us that this model is superior (and a good candidate for use in human experiments), each model has its advantages. In the next section, we turn to a more concrete assessment of sample quality.

\section{Experiment 1: Visual Turing Tests}

Though several objective heuristics exist for measuring the visual quality of generated images, because we are developing a generative model for use in experiments with human participants, it is more important that the generated images are of high visual quality from the subjective perspective of a human observer. This is particularly true for synthesized faces, which are notoriously susceptible to the ``uncanny valley'' effect \citep{mori1970uncanny}, where synthetic faces near the threshold of photorealism appear bothersome.

\begin{figure}[!h]
  \begin{center}
    \includegraphics[trim = 0mm 2mm 0mm 13mm, clip, width=1.0\linewidth]{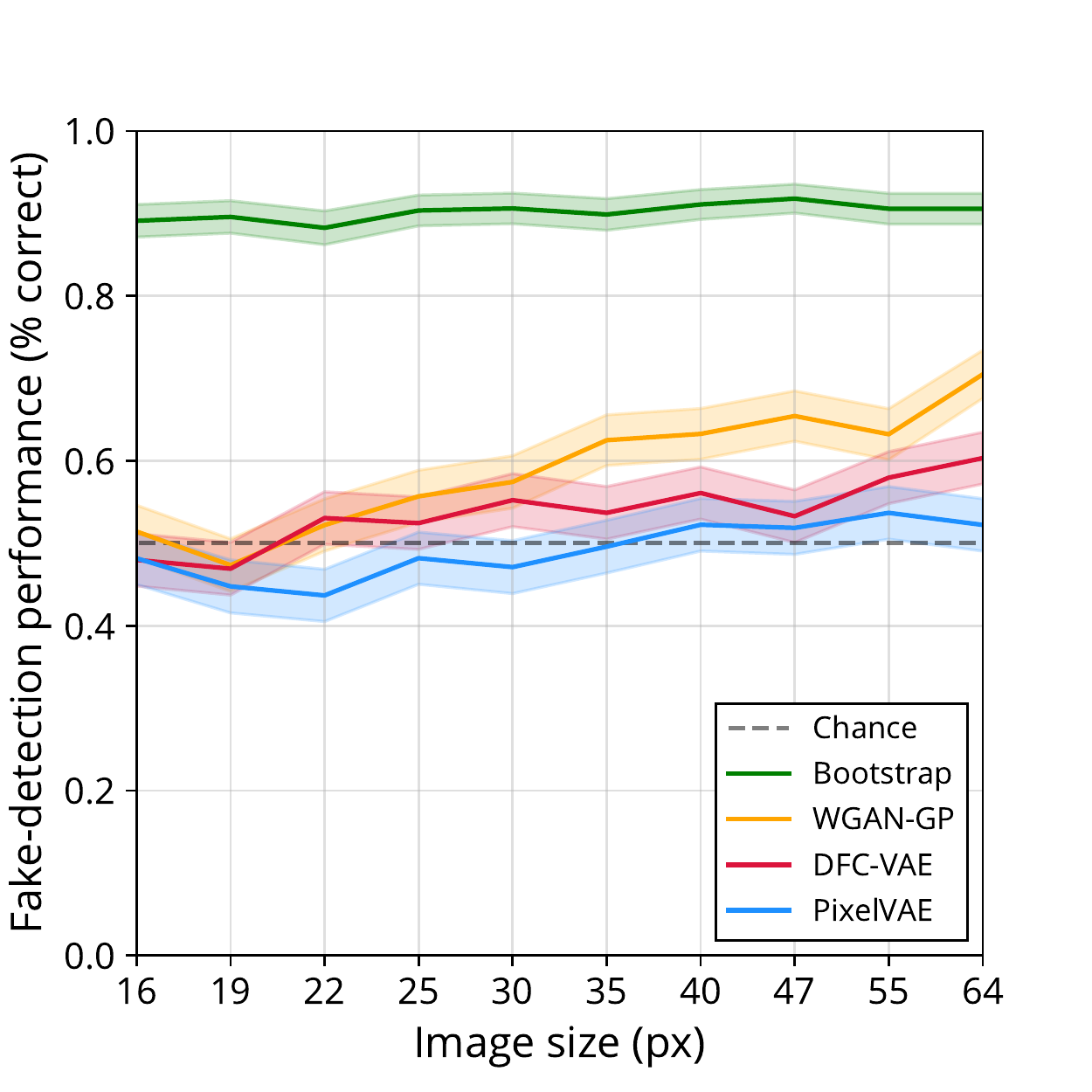}
  \vspace{-6mm}
  \caption{Psychophysical detection curves for each model. PixelVAE performs consistently better, and straddles the dotted line representing chance performance.}
  \label{turing_results}
  \end{center}
  \vspace{-5mm}
\end{figure}

To evaluate the degree to which we can traverse this valley, we use a visual Turing test for synthesized images that draws on methods from perceptual psychophysics to measure people's ability to distinguish real from synthesized photographs of faces, where the synthesized photographs are samples from each of the candidate models. This stands in contrast to two other forms of visual Turing test, one in which a human adjudicator determines which of two images was generated by human behavior \citep{lake2015human}, the other in which a visual recognition task is used to determine whether the observer performing the task is a human \citep{von2003captcha, geman2015visual}. Our test allows us to answer the question of which models can successfully fool humans and how this ability interacts with image size. The latter is important because successful generative models are often trained and tested with small images, e.g., $32\times32$ px, and it is unclear how many pixels are required for humans to make sensible judgments.

We collected such judgments from $250$ participants recruited from Amazon Mechanical Turk, each of whom performed $40$ trials, $10$ for each log-spaced size from $16\times16$ px to $64\times64$ px. On each trial, one image was selected from the training set and the other was a sample from one of the three candidate models. We also included a fourth candidate model as a control, one for which we know participants can easily distinguish samples from real photographs: a model that generates images by bootstrap resampling pixels from portraits in the Human{\ae} dataset. The results are plotted in Figure \ref{turing_results}. WGAN-GP performs worst, likely due to diverse artifacts, whereas PixelVAE consistently outperforms all other models in fooling humans. In fact, it stays near or below the line of chance for all image sizes, effectively traversing the uncanny valley for this domain. It is notable that near-perfect samples are rarely obtained in generative image models, outside of small domains such as MNIST \citep{lecun2010mnist} and SVHN \citep{netzer2011reading}. 

\section{Improving image quality\\ and exploring the latent space}

While our visual Turing tests show that our samples can fool human participants, they are still relatively small compared to the average comfortable stimulus sizes used in human experiments. This is an unfortunate feature of most modern generative image networks, which struggle to learn distributions over large sets of pixels. However, we note that the $64\times64$-pixel portraits generated by the PixelVAE carry enough information to convey identity \citep{bachmann1991identification}. This suggests our samples can be improved by a super-resolution network that enlarges the image, inventing plausible fine details while preserving identity.

\begin{figure*}[!h]
  \begin{center}
    \includegraphics[trim = 0mm 0mm 0mm 0mm, clip, width=1.00\textwidth]{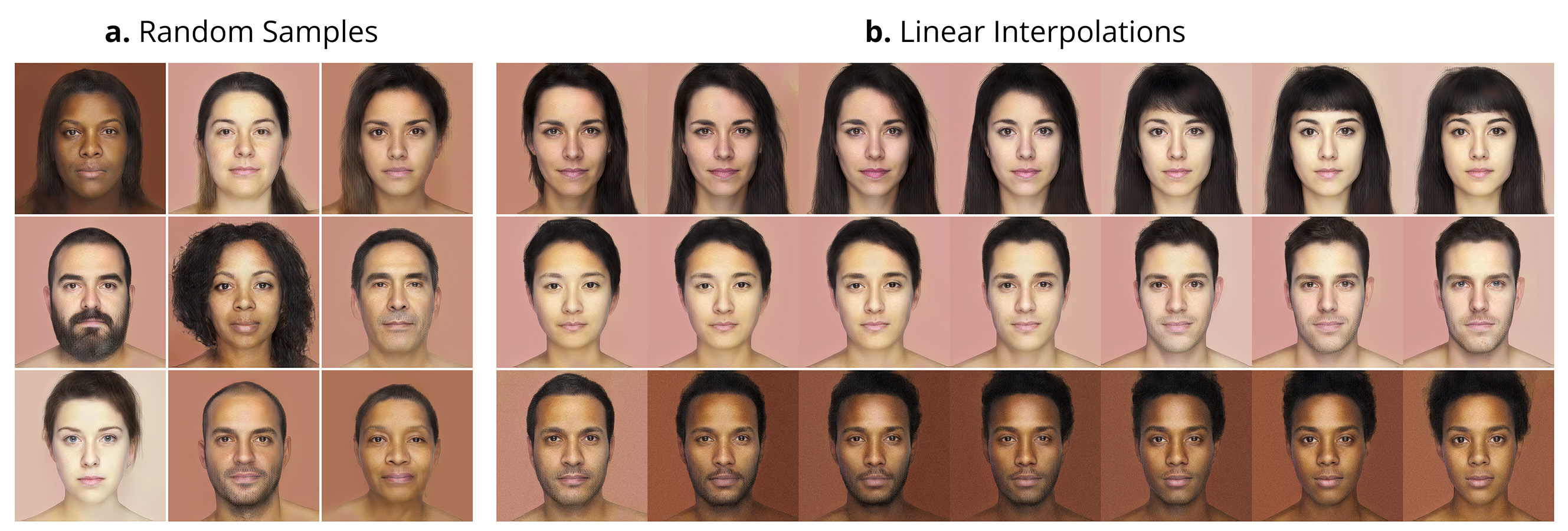}
  \vspace{-2mm}
  \caption{\textbf{a.} Decoded random samples drawn from the prior. \textbf{b.} Three decoded $7$-point linear interpolations between two random samples drawn from the prior.}
  \label{upsampled_samples}
  \end{center}
\end{figure*}

To do this, we use a generative adversarial network with an added content loss as our super-resolution upsampler network \citep{ledig2016photo}. We train this network to enlarge $64\times64$-pixel portrait inputs to $512\times512$-pixel outputs. Figure \ref{upsampled_samples}A shows enlarged PixelVAE samples. To our knowledge, these are among the highest quality synthesized facial identities produced by a deep neural network. Figure \ref{upsampled_samples}B demonstrates through enlargements of linear interpolations between random samples that identities and their composites are preserved, and only the quality and size of the image are improved.

We also tested for overfitting by randomly selecting 8 samples and finding their nearest neighbor in the training set, defined by pixelwise linear correlation. As evident in Figure \ref{overfitting}, samples and their nearest neighbors in the training set depict different identities.

\begin{figure*}[!ht]
  \begin{center}
    \includegraphics[trim = 0mm 0mm 0mm 0mm, clip, width=1.0\textwidth]{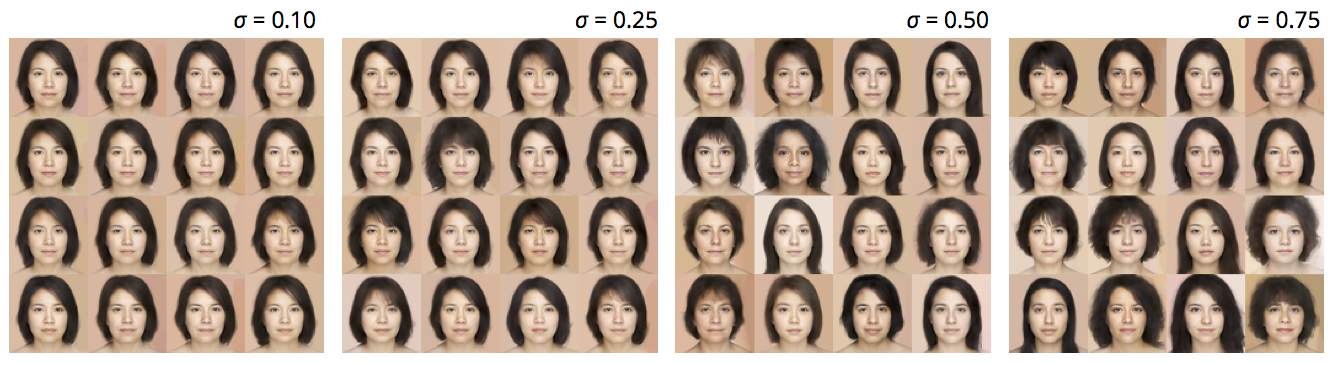}
  \vspace{-7mm}
  \caption{Latent noise perturbations at four levels of intensity.}
  \label{levels}
  \end{center}
  \vspace{-3mm}
\end{figure*}

\begin{figure}[!ht]
  \begin{center}
    \includegraphics[trim = 0mm 0mm 0mm 0mm, clip, width=0.25\textwidth]{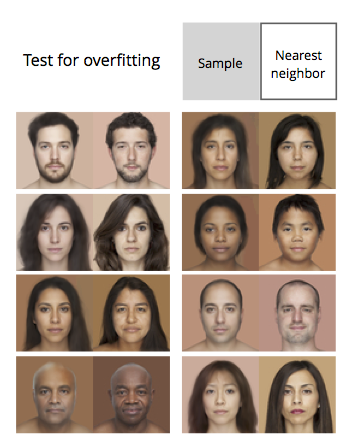}
  \vspace{-3mm}
  \caption{Samples and nearest neighbors in the training set.}
  \label{overfitting}
  \end{center}
  \vspace{-7mm}
\end{figure}

For our purposes, a useful machine representation is one that can be explored locally by humans and yield interesting variations on a concept. Figure \ref{levels} shows four levels of variation added to a base sample. This simple procedure allows us to propose small adjustments to facial features or hair, as well as large adjustments that dramatically alter identity along several meaningful dimensions. In the next section, we will use these properties to interact with human explorers.

\section{Experiment 2: Composite sketches via\\ interactive evolutionary computation}

A valuable but challenging application of generative models of human identity and appearance is the revelation of identities and appearances available only in a person's mind. For example, in police composite sketching, a trained artist with expertise in portraiture aims to translate witness testimony into a workable sketch that will enable law enforcement officials to recognize on sight the perpetrator of a crime  \citep{mancusi2010police}. Or when creating cover art for a book, an illustrator may wish to render a character's likeness using vague suggestions from the text.

Searching the latent space of a generative model of human identity provides a way to render these obscured identities. For example, software-based facial-composite tools used by some law enforcement agencies guide a witness through a series of sequential decisions about facial features to arrive at a well-formed composite sketch. When that generative model has been learned directly from a corpus of human identities, the model's latent space may reflect the underlying variation in a way that benefits the search. 

Various search and optimization algorithms are available, with some having proven particularly useful for human-in-the-loop search. Interactive evolutionary computation, a technique that inserts humans into various stages of an evolutionary process (e.g., as the fitness function), has been successfully used for human-in-the-loop search over digital designs and in other contexts \citep{takagi2001interactive}.

Here, we used a crowdsourced form of interactive evolutionary computation that leverages the human capacity to compare the relative resemblance of displayed portraits to a target identity. The particular evolutionary algorithm that we used comes from a family of so-called ``Natural Evolution Strategies'' that approximates stochastic gradient descent, making use of gradient information to efficiently find high-scoring pockets of the latent space \citep{spall1992multivariate}. The search begins by selecting a seed, a point $\theta_t$ in the latent space, typically the origin. On each round of the search, a lineup of $n$ portraits is generated by adding spherical Gaussian feature noise $\epsilon_{i}$ to the seed. A set of participants then ranks the $n$ portraits according to their resemblance to the target identity (e.g. ``a young boy with red hair'', ``Barack Obama''). Each portrait is assigned a score $F(\theta)$ equal to its average rank across all the participants. The next seed, $\theta_{t+1}$ is set to $\theta_t + \alpha \frac{1}{n\sigma}\sum_{i=1}^{n}F_{i}\epsilon_{i}$, where $\alpha$ is the learning rate. Thus, the next seed reflects the differences between high- and low-ranked examples to the extent that there is agreement between participants about the resemblance rankings.

Ten participants completed each round of the search. We found that interactive evolutionary computation over the latent space of our learned PixelVAE model efficiently found regions of the face space corresponding to target identities. Figure 8 shows the seeds and example images for 10 rounds of evolutionary search for three target categories.

\begin{figure*}[!h]
  \begin{center}
    \includegraphics[trim = 3mm 2mm 4mm 0mm, clip, width=1.0\textwidth]{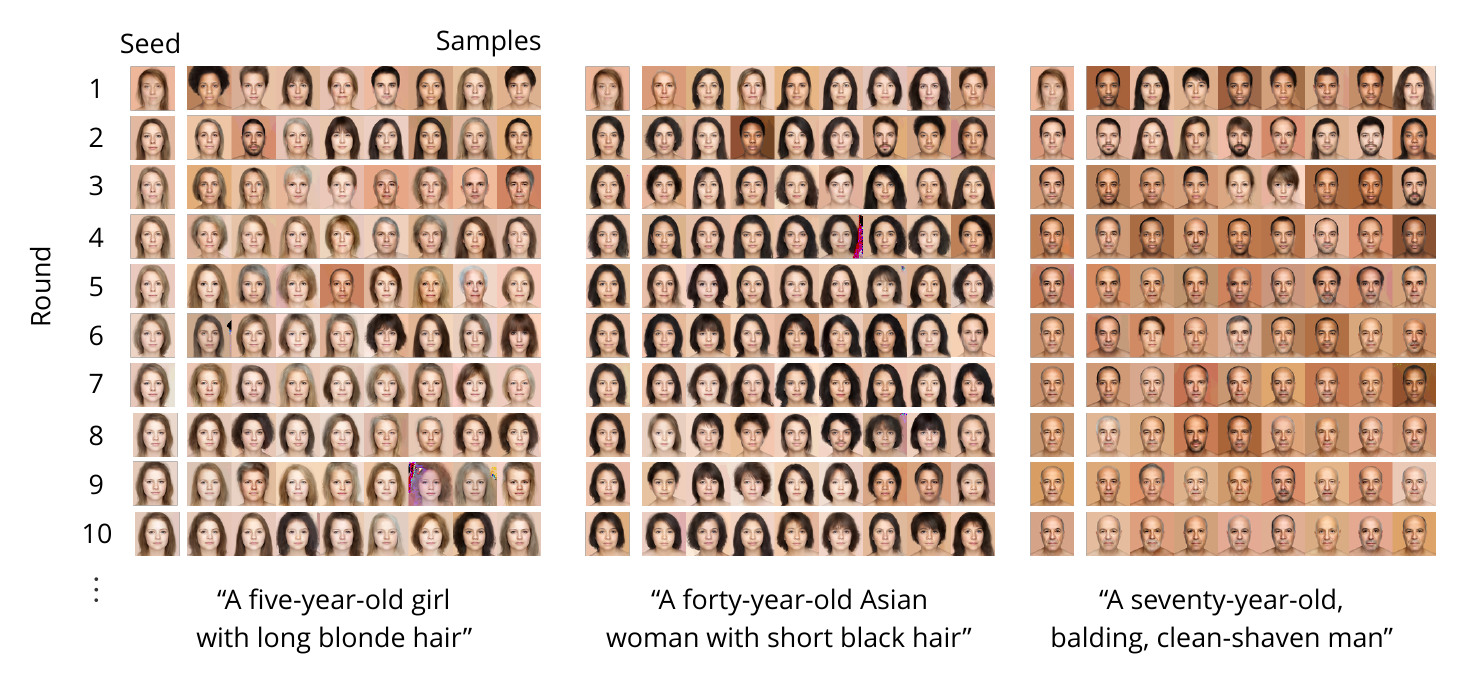}
  \vspace{-5mm}
  \caption{Three 10-round evolutionary searches through the latent space. The final seed for each set of trials represents a collective mental template for the verbal description at the bottom.}
  \label{searches}
  \end{center}
\vspace{-3mm}
\end{figure*}

\section{Discussion}

Here, we describe a finely tuned coupling between powerful generative models of human identity and appearance learned directly from data, and data that provides just the right variation essential to capturing the diversity of human identity and appearance. The samples produced by the models are interesting in their own right and rival the image quality of many state-of-the-art model--dataset pairs. However, we find these models most interesting in terms of what they tell us about humans, and how they can be applied to technology. Our hope is that the Human{\ae} dataset will inspire similar approaches, and that our models can be used as both empirical and practical tools. In the future, it will be worth exploring larger datasets, fine-tuning current models, testing new models, and finding new applications for interacting with humans and their own psychological representations of faces. Other questions remain. Can we approximately quantify the extent to which such spaces are universal? That is, do they contain every identity? To what extent do these space already align with human psychological representations of faces? And what are the limits to how much information about human identity and appearance can be extracted from a person's recollections?

\vspace{10pt}

\noindent \textbf{Acknowledgements.} This work was supported in part by the National Science Foundation (grant 1718550) and DARPA (cooperative agreement D17AC00004).

\renewcommand{\bibliographytypesize}{\small}
\bibliographystyle{apacite}
\setlength{\bibleftmargin}{.125in}
\setlength{\bibindent}{-\bibleftmargin}
\bibliography{citations}

\end{document}